\def\paperID{12032}
\def\confName{CVPR}
\def\confYear{2025}

\def\paperTitle{Adversarial Training in Low-Label Regimes with Margin-Based Interpolation}

\def\authorBlock{
    Tian Ye \qquad
    Rajgopal Kannan \qquad
    Viktor Prasanna \\
    University of Southern California, DEVCOM Army Research Office \\
    {\tt\small \{tye69227, prasanna\}@usc.edu, rajgopal.kannan.civ@army.mil}
}

\newif\ifreview 
\newif\ifarxiv \newcommand{\arxiv}{\arxivtrue}
\newif\ifcamera 
\newif\ifrebuttal 

\arxiv 

\pdfoutput=1
\documentclass[10pt,twocolumn,letterpaper]{article}
\ifreview \usepackage[review]{cvpr} \fi
\ifarxiv \usepackage[pagenumbers]{cvpr} \fi
\ifrebuttal \usepackage[rebuttal]{cvpr} \fi
\ifcamera \usepackage{cvpr} \fi

\usepackage{graphicx}	
\usepackage{amsmath}	
\usepackage{amssymb}	
\usepackage{booktabs}
\usepackage{times}
\usepackage{microtype}
\usepackage{epsfig}
\usepackage{caption}
\usepackage{float}
\usepackage{placeins}
\usepackage{color, colortbl}
\usepackage{stfloats}
\usepackage{enumitem}
\usepackage{tabularx}
\usepackage{xstring}
\usepackage{multirow}
\usepackage{xspace}
\usepackage{url}
\usepackage{subcaption}
\usepackage{xcolor}
\usepackage[hang,flushmargin]{footmisc}

\usepackage{algorithm}
\usepackage{algpseudocode}

\usepackage{dsfont}

\usepackage{enumitem}
\usepackage{comment}

\usepackage{threeparttable}

\usepackage{pifont}
\newcommand{\cmark}{\ding{51}}%
\newcommand{\xmark}{\ding{55}}%

\ifcamera \usepackage[accsupp]{axessibility} \fi

% ...

\ifarxiv  \fi

\newcommand{\Reviewer}[1]{{%
    \textbf{%
        \ifstrequal{#1}{1}{\textcolor{red}{R#1}}{%
        \ifstrequal{#1}{2}{\textcolor{blue}{R#1}}{%
        \ifstrequal{#1}{3}{\textcolor{magenta}{R#1}}{%
        \ifstrequal{#1}{4}{\textcolor{teal}{R#1}}{%
                           \textcolor{cyan}{R#1}%
        }}}}%
    }%
}}

\newcommand{\calX}{{\mathcal{X}}}
\newcommand{\calY}{{\mathcal{Y}}}

\newcommand{\calA}{{\mathcal{A}}}
\newcommand{\calB}{{\mathcal{B}}}

\newcommand{\R}{\mathbb{R}}

\newcommand{\KL}{D_\text{KL}}
\newcommand{\CE}{\ell_\text{ce}}

\newcommand{\rbr}[1]{\left(#1\right)}

\newcommand{\norm}[1]{\left\| #1\right\|}

\usepackage{xr}

\makeatletter
\newcommand*{\addFileDependency}[1]{
  \typeout{(#1)}
  \@addtofilelist{#1}
  \IfFileExists{#1}{}{\typeout{No file #1.}}
}

\makeatother
\newcommand*{\myexternaldocument}[1]{
    \externaldocument{#1}
    \addFileDependency{#1.tex}
    \addFileDependency{#1.aux}
}

\definecolor{cvprblue}{rgb}{0.21,0.49,0.74}
\usepackage[pagebackref,breaklinks,colorlinks,allcolors=cvprblue]{hyperref}
\usepackage[capitalize]{cleveref}
\crefname{section}{Sec.}{Secs.}
\crefname{table}{Table}{Tables}
\crefname{figure}{Fig.}{Figs.}

\ifarxiv \crefname{appendix}{App.}{Apps.}
\else \crefname{appendix}{Suppl.}{Suppls.} \fi

\unless\ifarxiv \myexternaldocument{_supplementary} \fi

\begin{document}
\title{\paperTitle}
\author{\authorBlock}
\maketitle

\renewcommand{\thefootnote}{} 
\footnotetext{This work is supported by the DEVCOM Army Research Office (ARO) under grant W911NF2220159 and W911NF2320186.\\
\textbf{Distribution Statement A:} Approved for public release. Distribution is unlimited. }
\renewcommand{\thefootnote}{\arabic{footnote}} 

\begin{abstract}
Adversarial training has emerged as an effective approach to train robust neural network models that are resistant to adversarial attacks, even in low-label regimes where labeled data is scarce. In this paper, we introduce a novel semi-supervised adversarial training approach that enhances both robustness and natural accuracy by generating effective adversarial examples. Our method begins by applying linear interpolation between clean and adversarial examples to create interpolated adversarial examples that cross decision boundaries by a controlled margin. This sample-aware strategy tailors adversarial examples to the characteristics of each data point, enabling the model to learn from the most informative perturbations. Additionally, we propose a global epsilon scheduling strategy that progressively adjusts the upper bound of perturbation strengths during training. The combination of these strategies allows the model to develop increasingly complex decision boundaries with better robustness and natural accuracy. Empirical evaluations show that our approach effectively enhances performance against various adversarial attacks, such as PGD and AutoAttack.
\end{abstract}

\section{Introduction}
\label{sec:intro}
While widely applied in computer vision tasks, deep neural networks have been shown vulnerable to adversarial attacks~\cite{fgsm,i-fgsm,pgd,autoattack,ilyas}. During inference time, attackers can drastically change the output of a neural network by manipulating the input data with perturbations imperceptible to human eyes. 
Such vulnerability of neural networks can be damaging in critical vision applications, such as autonomous driving~\cite{driving1,driving2}, medical imaging~\cite{medical1}, and surveillance systems~\cite{surveillance1}.
As a defense strategy, \textbf{adversarial training}~\cite{pgd,trades,arow,gairat,hat,mail,marts,mma} is an effective method to improve the robustness of neural networks against adversarial attacks. For each clean sample $x$ from the training dataset, an adversarial example $x^\text{adv}$ within the $\epsilon$-ball around $x$ is generated and utilized to train the neural network models, where $\epsilon$ is the \textbf{perturbation strength} that controls the distance between $x^\text{adv}$ and $x$ in terms of $\ell_p$-norm. This allows the models to learn the features in adversarial examples and thus show robustness against similar attacks. While adversarial training can effectively improve the \textbf{robustness} of neural networks, it often comes at the cost of lower \textbf{natural accuracy} on clean data~\cite{tradeoff,trades}. 
Balancing robustness with natural accuracy remains a key consideration in deploying robust machine learning models in real-world applications.

A challenge in training robust models is the need for large amount of labeled training data. In real-world scenarios, acquiring unlabeled image data is typically easier and more cost-effective than obtaining labeled data. To address this challenge and leverage unlabeled data effectively, \textbf{semi-supervised adversarial training}~\cite{vat,rst,uat,srstawr} has gained attention as a method for training robust models in low-label regimes, where labeled data is scarce. 

In low-label regimes, datasets exhibit diversity across data points in two key aspects. First, there is variability in vulnerability to adversarial attacks: Some data points lie close to decision boundaries, allowing adversarial examples to be found within very small $\epsilon$-balls, while others positioned farther from the decision boundaries require stronger perturbations to find adversarial examples. Second, there is a disparity in label availability: Some data points in the training set have ground truth labels, while others are unlabeled and thus lack class information. Given this inherent diversity, enhancing the robustness and natural accuracy of semi-supervised adversarial training requires methods that can effectively leverage all data points, particularly the large majority of unlabeled data, while accounting for their individual characteristics. This observation motivates the development of adaptive techniques that can better exploit datasets in low-label regimes, resulting in more robust and accurate models.

\begin{figure}
    \centering
    \includegraphics[width=0.95\linewidth]{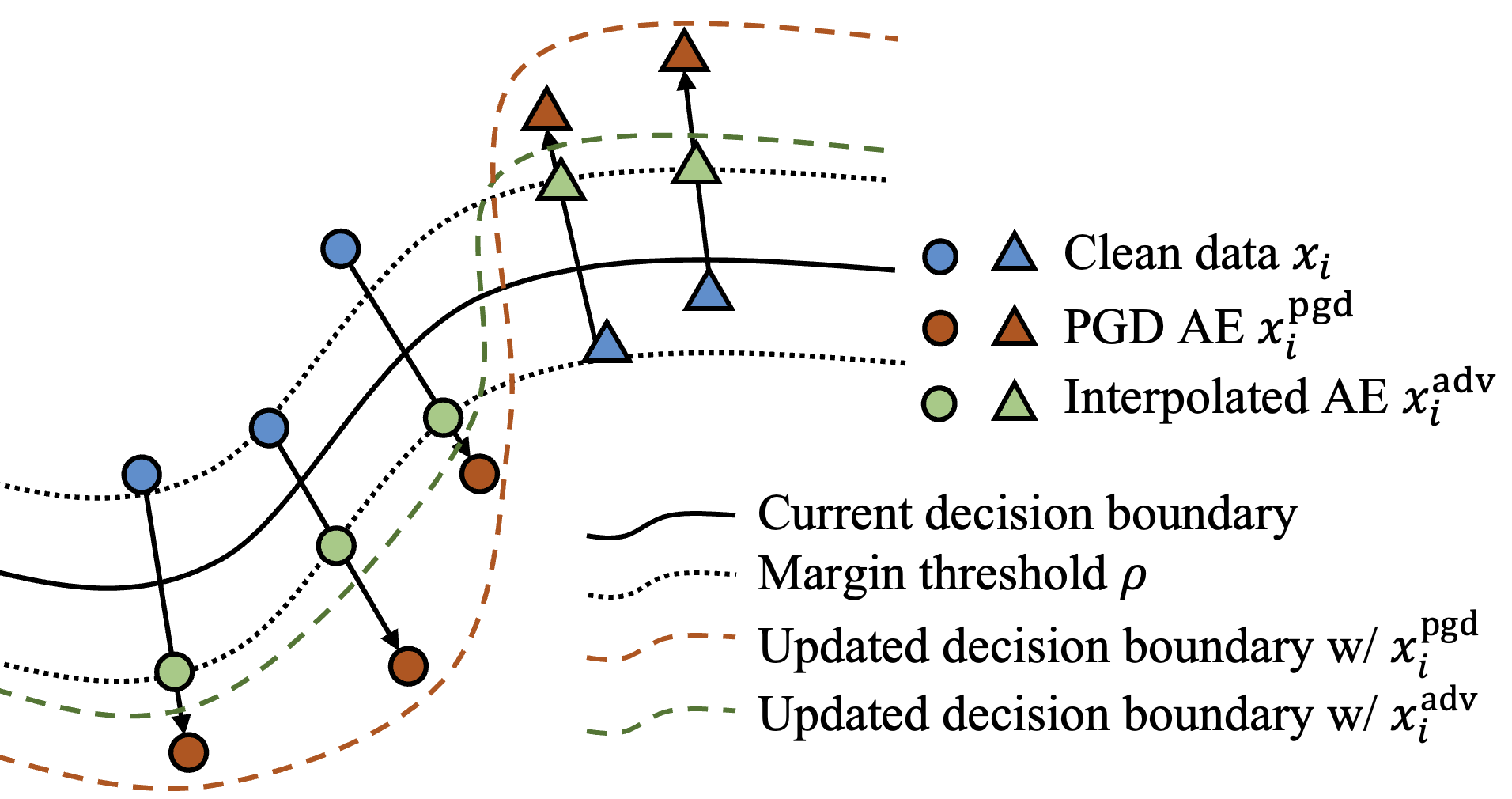}
    \caption{Illustration of updating the decision boundary of a classifier with interpolated adversarial examples.}
    \label{fig:illustrate}
\end{figure}

In this work, we introduce a novel semi-supervised adversarial training algorithm designed for low-label regimes, where only a small fraction of the training set is labeled, and the majority remains unlabeled. 
Our algorithm uses an adaptive approach that adjusts perturbation strength for each individual data point. Specifically, for each data point, whether labeled or unlabeled, an adversarial example is generated within its $\epsilon$-ball. We then create \textbf{interpolated adversarial examples} by interpolating between the clean data point and its corresponding adversarial example, as illustrated in~\cref{fig:illustrate}. These interpolated adversarial examples are carefully controlled to cross the decision boundaries by a small margin, with a margin function defined for both labeled data and unlabeled data. This sample-aware approach allows us to effectively utilize all data points and enables neural network models to develop more nuanced decision boundaries, enhancing their generalization to unseen attacked data while maintaining high natural accuracy on clean data.

Additionally, we propose \textbf{global epsilon scheduling} that controls the overall perturbation strength throughout the training process. Specifically, we design a strategy that incrementally increases the perturbation strength $\epsilon$ (before interpolation) across training epochs. This strategy allows the model to start with easier adversarial examples and gradually progress to more challenging ones. 
This enables more comprehensive exploration of adversarial examples and leads to improved robustness.
The combination of global epsilon scheduling with interpolated adversarial examples synergistically enhances robust model training, where the former sets a global upper bound on perturbation strengths across epochs, and the latter adjusts perturbation strengths within this upper bound for individual data points. While our approach shares some similarities with existing works, we outline key differences in~\cref{sec:compare-with-prior}.

Our contributions are summarized as follows:
\begin{itemize}
    \item 
    We propose a sample-aware method that uses interpolation between clean and adversarial examples. Our method controls the margin by which the interpolated adversarial examples cross the decision boundaries, enabling the model to learn from adversarial attacks with appropriate perturbation strengths and thus effectively exploit the training set in low-label regimes where labeled data is scarce.
    \item We propose a global epsilon scheduling strategy that controls the global upper bound of perturbation strengths, starting from low levels and gradually increasing as the training proceeds. This progressive approach helps the model to iterate their decision boundaries that eventually generalize better to unseen data. 
    \item 
    We introduce Semi-Supervised Adversarial Training with Margin-Based Interpolation (SSAT-MBI), a novel algorithm that integrates our two proposed techniques. SSAT-MBI can also be integrated with existing semi-supervised adversarial training approaches. Extensive experiments on multiple datasets for image classification demonstrate that our method significantly enhances both robustness and natural accuracy compared to baseline approaches.
\end{itemize}

The rest of this paper is organized as follows. \Cref{sec:related} defines the problem setting and briefly reviews the related works. \Cref{sec:method} presents our approach of semi-supervised adversarial training. 
We present experimental results to demonstrate the effectiveness of our approach in \Cref{sec:experiments} and conclude this paper in~\Cref{sec:conclusion}.

\bigskip

\section{Preliminaries and Related Work}
\label{sec:related}

\subsection{Notations }
Considering a classification problem with $C$ classes, let $\calX\subset \R^d$ be the input space and $\calY= \{1,2,...,C\}$ be the labels. For low-label regimes where labeled data is scarce, let $D_L=\{(x_1,y_1),...,(x_n,y_n)| (x_i,y_i)\in \calX\times\calY\}$ be a set of labeled data points and 
$D_U=\{x_{n+1},...,x_{n+m}| x_{n+j}\in \calX\}$ be a set of unlabeled data points. The size of labeled data is much smaller than unlabeled data, i.e., $n\ll m$.

Let $f_\theta:\calX\rightarrow \R^C$ be a classifier with parameters $\theta$, and let $p_\theta(\cdot| x)=\sigma(f_\theta(x))\in[0,1]^C$ be the predicted likelihood that $x$ belongs to each class, where $\sigma(\cdot)$ is the softmax function.
Let $\calB_p(x,\epsilon)=\{x'\in\calX:\norm{x'-x}_p\le\epsilon\}$ be the $\epsilon$-ball around $x$, where $\norm{\cdot}_p$ is the $\ell_p$-norm. 

\subsection{Adversarial Attack}
\label{sec:attack}

To attack a trained classifier $f_\theta$ at inference time, an adversarial attack can manipulate a clean data point $x$ with an imperceptible perturbation $\delta\in \calB_p(0,\epsilon)$ and generate an adversarial example $x^\text{adv}=x+\delta$, such that $x^\text{adv}$ is misclassified by $f_\theta$, i.e., $\arg\max_{k\in\calY}p_\theta(k|x^\text{adv})\ne y$, where $y$ is the ground truth class of $x$. 
Let $\CE$ be the cross-entropy loss function. Then an adversarial attack can be implemented as an optimization problem:
\begin{equation}
\delta = \arg\max_{\delta \in \calB_p(0,\epsilon)} \CE(f_\theta(x+\delta), y).    
\end{equation}
A popular approach to solve this optimization is the Projected Gradient Descent (PGD-$T$)~\cite{pgd}, which updates a proposed perturbation iteratively in $T$ steps:
\begin{equation}
\delta^{(t+1)}=\Pi_{\calB_p(0,\epsilon)}\left[\delta^{(t)}+\eta\cdot\text{sgn}\rbr{\nabla_x\CE(f_\theta(x+\delta^{(t)}), y)}\right],
\end{equation}
where $\delta^{(0)}$ is randomly initialized in $\calB_p(0,\epsilon)$, and $x^\text{adv}=x+\delta^{(T)}$. The operator $\Pi_{\calB_p(0,\epsilon)}$ projects the perturbation to the $\epsilon$-ball, and $\eta$ is the step size.

\subsection{Semi-Supervised Adversarial Training}

Adversarial training~\cite{pgd,trades,arow,gairat,hat,mail,marts,mma} can effectively improve the robustness of a classifier by incorporating adversarial examples into the training process. During training time, each clean sample from the training set is perturbed by adversarial attacks (e.g., PGD-10) to generate adversarial examples (referred to as the \textit{inner maximization} phase). 
The adversarial examples are then used to train the classifier (referred to as the \textit{outer minimization} phase), which can thus have robustness against similar attacks during inference time.

For a labeled data point $(x_i,y_i)\in D_L$, the inner maximization that generates adversarial examples can be implemented by PGD-$T$ as described in Section~\ref{sec:attack}.
For an unlabeled data point $x_i\in D_U$, the adversarial example can be alternatively represented as 
\begin{equation}
\label{eq:inner2}
x_i^\text{adv} = \arg\max_{x' \in \calB_p(x_i,\epsilon)} \CE(f_\theta(x'), \tilde{y}_i),    
\end{equation}
where $\tilde{y}_i=p_{\theta_T}(\cdot|x_i)\in\R^C$ be the pseudo-label with $\theta_T$ being a model parameter trained by a (non-robust) semi-supervised learning, e.g., FixMatch~\cite{fixmatch}.
Alternatively, for both labeled and unlabeled data, KL divergence can also be used to replace $\CE$, in which case the adversarial example is
\begin{equation}
\label{eq:inner3}
x_i^\text{adv}=\arg\max_{x' \in \calB_p(x_i,\epsilon)} \KL(p_\theta(\cdot|x_i)\parallel p_\theta(\cdot|x')).    
\end{equation}
Both variants of the inner maximization can be solved by PGD-$T$.

With adversarial examples generated, existing semi-supervised adversarial training methods have different designs on the outer minimization. Robust Self-Training (RST)~\cite{rst} minimizes the following objective:
\begin{equation}
\label{eq:rst}
\begin{aligned}
\min_\theta \frac{1}{n+m} \sum_{i=1}^{n+m} &\{\CE(f_\theta(x_i),y_i \text{ or } \tilde{y_i})\\
&+\lambda\cdot\KL(p_\theta(\cdot|x_i)\parallel p_\theta(\cdot|x_i^\text{adv}))\}.    
\end{aligned}
\end{equation}
where $\lambda > 0$ is a hyperparameter balancing the two loss terms. This combines the loss on clean data points, and a \textit{consistency loss} to encourage the classifier to show similar outputs for clean and adversarial examples.
Unsupervised Adversarial Training (UAT++)~\cite{uat} minimizes a similar objective as RST:
\begin{equation}
\label{eq:uat}
\begin{aligned}
\min_\theta \frac{1}{n+m} \sum_{i=1}^{n+m} &\{\CE(f_\theta(x_i^\text{adv}),y_i \text{ or } \tilde{y_i})\\
&+\lambda\cdot\KL(p_{\hat{\theta}}(\cdot|x_i)\parallel p_\theta(\cdot|x_i^\text{adv}))\},    
\end{aligned}
\end{equation}
where $\hat{\theta}$ is a fixed copy of $\theta$. 
A more recent work proposed 
Semi-Supervised Robust Self-Training via Adaptively Weighted Regularization (SRST-AWR)~\cite{srstawr} where a weight factor is introduced to the consistency loss of RST: 
\begin{equation}
\label{eq:srst1}
    \KL(p_{\theta}(\cdot|x_i) \parallel p_{\theta}(\cdot|x^\text{adv}_i)) \cdot w_{\theta}(x_i)
\end{equation}
\begin{equation}
\label{eq:srst2}
w_{\theta}(x_i) = \frac{1}{2}\sum_{c=1}^{C} \tilde{y}_{i,c} p_\theta(c|x_i)+\frac{1}{2}\sum_{c=1}^{C} \tilde{y}_{i,c}\left(1-p_\theta(c|x_i^\text{adv})\right),  
\end{equation}
which imposes more weights to the consistency loss when the prediction on clean data $x_i$ is better correlated to the pseudo-label $\tilde{y}_i$ (i.e., $f_\theta$ has good performance on $x_i$) and the prediction on adversarial example $x_i^\text{adv}$ is less correlated to $\tilde{y}_i$ (i.e., $f_\theta$ needs to improve its robustness on $x_i^\text{adv}$). 

Different from these existing methods that primarily focus on designing objectives for the outer minimization phase, our method enhances semi-supervised adversarial training at the inner maximization phase where we select appropriate adversarial examples. As such, our method can be seamlessly integrated with these methods, as described in~\cref{sec:half}.

\bigskip

\section{Methodology}
\label{sec:method}

\begin{figure*}[htbp]
  \centering
    \begin{subfigure}[b]{0.495\textwidth}
    \centering
    \includegraphics[height=5.8cm]{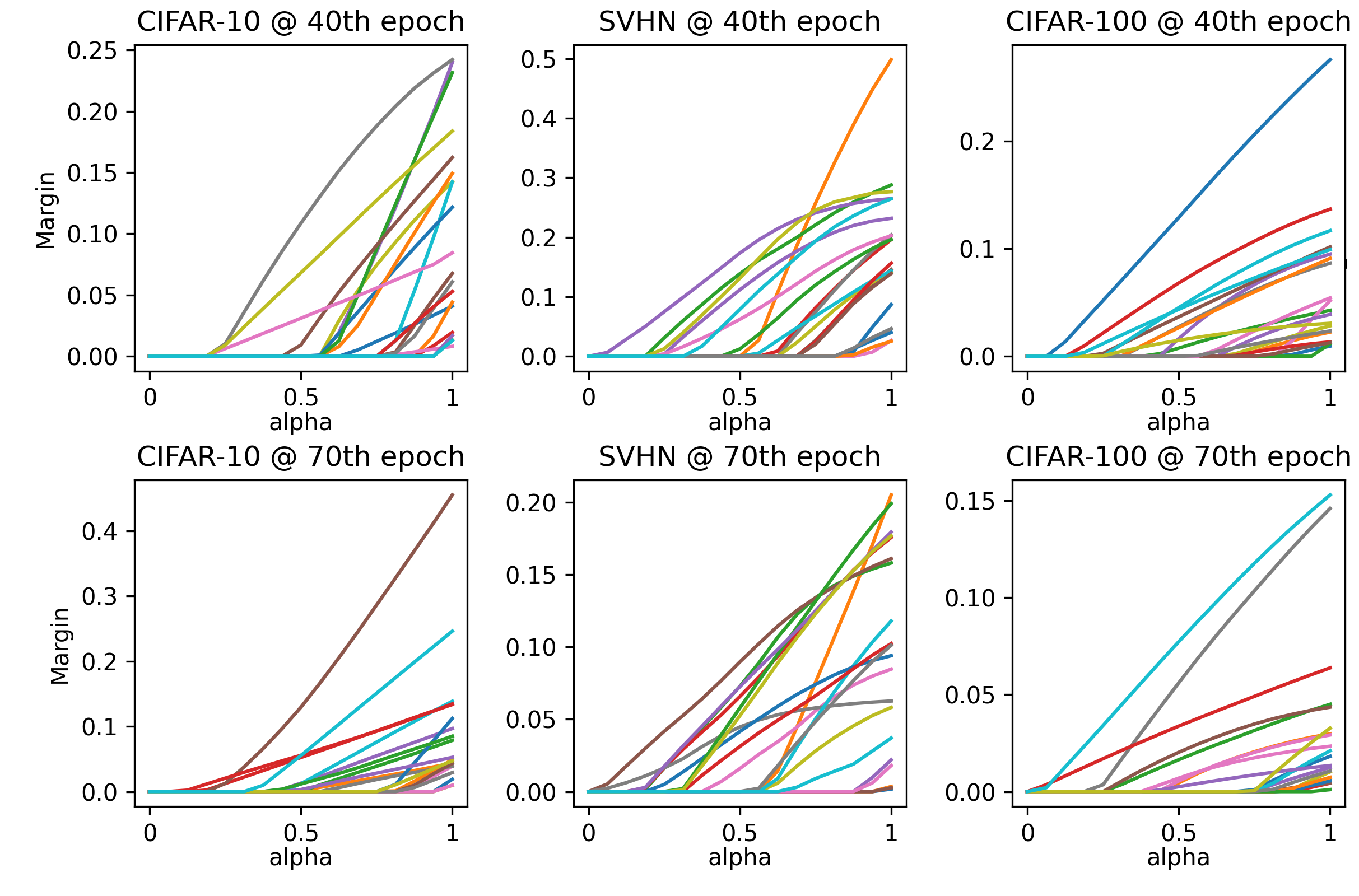}
    \caption{}
    \label{fig:asm1}
    \end{subfigure}
    \hfill
    \begin{subfigure}[b]{0.495\textwidth}
    \centering
    \includegraphics[height=5.8cm]{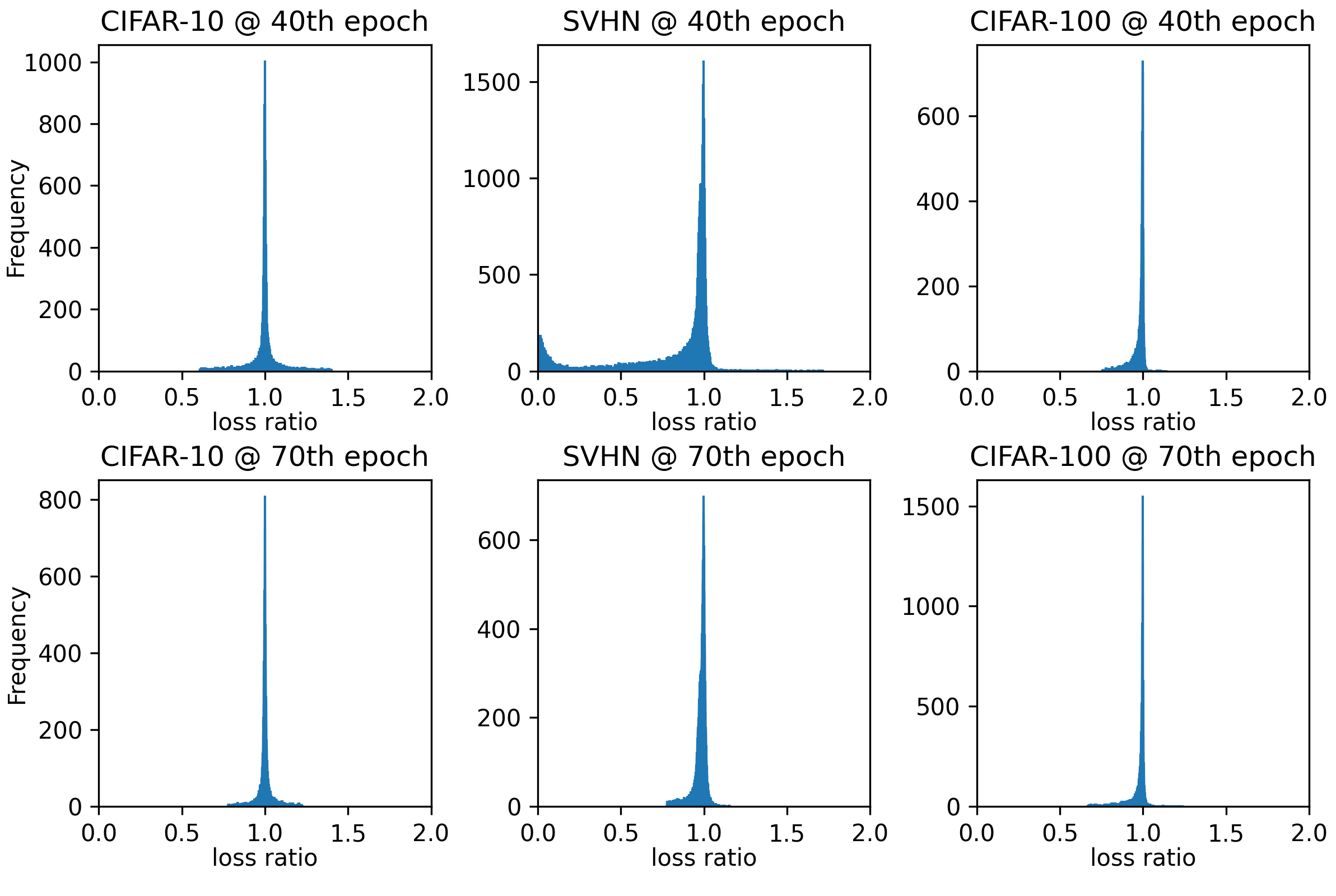} 
    \caption{}
    \label{fig:asm2}
  \end{subfigure}
  \caption{Empirical supports for Assumption 1 in~\cref{sec:intp} and Assumption 2 in \cref{sec:insights}, based on three different datasets (CIFAR-10, SVHN, CIFAR-100) at epochs 40 and 70.  \textbf{(a)} The value of margin function $d(\alpha; x_i,x_i^\text{pgd})$ for 20 randomly sampled training data points as $\alpha$ varies from 0 to 1. \textbf{(b)} Histogram of the loss ratio  $\CE(f_\theta(x_i^\text{adv}(\hat{\alpha})),\tilde{y}_i)/\CE(f_\theta(\widehat{x_i}^\text{pgd}),\tilde{y}_i)$ as defined in Assumption 2 in \cref{sec:insights}.}
  \label{fig:assumption}
\end{figure*}

This section presents our proposed methodology in detail. In~\cref{sec:saps}, we introduce our proposed technique of interpolated adversarial examples. In~\cref{sec:dynamic-epsilon}, we describe our global epsilon scheduling strategy. We then integrate these techniques and describe our Semi-Supervised Adversarial Training with Margin-Based Interpolation (SSAT-MBI) algorithm in~\cref{sec:half}. Finally, we compare our proposed techniques with prior works in~\cref{sec:compare-with-prior}.

\subsection{Sample-Aware Perturbation Strength Based on Margin}
\label{sec:saps}

In existing semi-supervised adversarial training methods~\cite{rst,uat,srstawr}, a fixed perturbation strength $\epsilon$ is typically used to generate adversarial examples, which fails to account for the inherent diversity among data points as discussed in~\cref{sec:intro}. Our method addresses this by regulating the margin by which adversarial examples cross the decision boundary, taking two key aspects of data diversity into account: (1) Data points closer to the decision boundary are assigned weaker perturbations, while those farther away receive stronger perturbations, ensuring that the margin is appropriate for each data point. (2) For labeled data, the margin is computed using ground truth labels, whereas for unlabeled data, soft pseudo-labels are used, with the confidence of these labels influencing the margin. By adapting the perturbation strength for each data point based on the margin, our method generates more informative adversarial examples, thereby improving both the robustness and natural accuracy of the models.

\subsubsection{Margin of Adversarial Examples for Unlabeled Data}
To achieve this, we define a margin function as follows.
For data point $x_i\in\calX$ with its adversarial example $x_i^\text{adv}\in\calB(x_i,\epsilon)$, a score $s(x_i^\text{adv})$ that represents the model's prediction on its class is defined as
\begin{equation}
    \label{eq:score}s(x_i^\text{adv})=\sigma(f_\theta(x_i^\text{adv})/\tau) \in\mathbb{R}^C,
\end{equation}
where $\sigma(\cdot)$ is the softmax function, and temperature $\tau$ controls the smoothness of the distribution of scores across classes and allows us to adjust its sensitivity.
For a labeled data point $(x_i,y_i)\in D_L$ and its adversarial example $x_i^\text{adv}$, the margin is defined as 
\begin{equation}
\label{eq:d_labeled}
    d(x_i^\text{adv})=\max_{k\in\calY}[s(x_i^\text{adv})]_k - [s(x_i^\text{adv})]_{y_i},    
\end{equation}
which is the difference between the score of the most likely class and the score of the ground truth class.
For an unlabeled data point $x_i\in D_U$ with pseudo-label $\tilde{y}_i=p_{\theta_T}(\cdot|x_i)\in \R^C$ where $\theta_T$ is teacher parameters trained by a (non-robust) semi-supervised learning algorithm over $D_L\cup D_U$, the margin is defined as
\begin{equation}
\label{eq:d_unlabeled}
    d(x_i^\text{adv})=\max_{k\in\calY}[s(x_i^\text{adv})]_k-\sum_{j=1}^C\tilde{y}_{i,j}[s(x_i^\text{adv})]_j.    
\end{equation}
\Cref{eq:d_unlabeled} can be seen as a ``soft" version of \cref{eq:d_labeled} when the labels are unavailable and probabilistic pseudo-labels are used instead. For brevity, we can rewrite the label $y_i\in\calY$ for labeled data as a one-hot vector $\tilde{y}_i\in \R^C$, so that \cref{eq:d_unlabeled} applies to both labeled and unlabeled data.

\subsubsection{Margin-Based Interpolation}
\label{sec:intp}

With the margin defined for any data point $x_i\in\calX$, we consider a method based on interpolation. Specifically, let $x_i$ be a clean data point and $x_i^\text{pgd}$ the adversarial example generated by PGD within the $\epsilon_\text{max}$-ball of $x_i$. Here, $\epsilon_\text{max}$ represents the upper bound of perturbation strength, which will be determined by the global epsilon scheduling strategy described in \cref{sec:dynamic-epsilon}. With a factor $\alpha\in [0,1]$, we define $x_i^\text{adv}$ as the linear interpolation between $x_i$ and $x_i^\text{pgd}$, i.e.,
\begin{equation}
    x_i^\text{adv}(\alpha)=\alpha\cdot x_i^\text{pgd}+(1-\alpha)\cdot x_i,
\end{equation}
referred to as an \textbf{interpolated adversarial example}.

With a slight abuse of notation, we can express the margin for the interpolated adversarial example as $d(\alpha;x_i, x_i^\text{pgd})$, which is a function of $\alpha$ with $x_i$ and $x_i^\text{pgd}$ fixed. Our goal is to find a maximum $\hat{\alpha}$ such that $d(\hat{\alpha};x_i, x_i^\text{pgd})\le \rho$, where $\rho>0$ is a hyperparameter for the threshold to control the margin of adversarial examples. This can be approximated using a binary search, as detailed in Line 9-18 of \cref{algo:train}. 
Note that computing the interpolated adversarial example incurs only a slight computational overhead. It requires $K$ forward propagations through the model $f_\theta$ (Line 13 of Algorithm~\ref{algo:train}), without the need for additional time-consuming PGD attacks beyond the one performed in Line 8.

To justify the effectiveness of binary search, we make the following assumption:
\begin{enumerate}[leftmargin=*, label={}]
    \item \textbf{Assumption 1.} For most data points where $x_i$ is correctly classified and $x_i^\text{pgd}$ is misclassified by $f_\theta$, the function $d(\alpha;x_i, x_i^\text{pgd})$ is non-decreasing as $\alpha$ increases from 0 to 1. 
\end{enumerate}
Assumption 1 allows us to use binary search to efficiently find $\hat{\alpha}$ such that $d(\hat{\alpha};x_i, x_i^\text{pgd})\approx \rho$.
This assumption is reasonable because $x_i^\text{pgd}$ effectively pushes the model's prediction further away from the true class. Consequently, the margin function $d(\alpha;x_i, x_i^\text{pgd})$, which represents the difference between the model's highest predicted score and the score for the true class, tends to increase as the interpolated adversarial example shifts from $x_i$ to $x_i^\text{pgd}$. Although this behavior may not be strictly monotonic due to the complex nature of neural network decision boundaries, we can empirically demonstrate the overall trend of increasing $d$, as shown in \cref{fig:asm1}.

\subsubsection{Interpretation of Interpolated Adversarial Examples}
\label{sec:insights}

The interpolated adversarial examples can be seen as an approximation of the following optimization problem. 
For any data point $x_i\in\calX$, the sample-aware perturbation strength can be formulated as the maximum of $\epsilon_i$ with the constraints
\begin{equation}
\left\{
\begin{aligned}
&\epsilon_i\le\epsilon_\text{max},\\
    &d(x_i^\text{adv})\le\rho, \\
    &\quad \text{where } x_i^\text{adv}=\arg\max_{x'\in\calB_p(x_i,\epsilon_i)}\CE(f_\theta(x'),\tilde{y}_i). \\
\end{aligned}
\right.
\label{eq:aps}
\end{equation}
As the relationship between $\epsilon_i$ and $d(x_i^\text{adv})$ is dependent on the current status of the model $f_\theta$, the process of finding $\epsilon_i$ needs to repeat for each data sample in every training epoch. Thus, it is necessary to efficiently find the maximum $\epsilon_i$ that satisfies \cref{eq:aps} with low computational overhead. 
However, generating $x_i^\text{adv}$ requires an iterative process (e.g., PGD), making it computationally expensive to explore different $\epsilon_i$ values and find the optimal one.

With the following assumption, the interpolation between clean data point $x_i$ and adversarial example $x_i^\text{pgd}$ produced by PGD, as described in \cref{sec:intp}, provides an approximated solution. 

\begin{enumerate}[leftmargin=*, label={}]
    \item \textbf{Assumption 2.} 
    Let $\hat{\epsilon}=\norm{x_i^\text{adv}(\hat{\alpha})-x_i}$, i.e., the perturbation strength of the interpolated adversarial example when $\alpha=\hat{\alpha}$. Let $\widehat{x_i}^\text{pgd}$ be the adversarial example generated by PGD within the $\hat{\epsilon}$-ball of $x_i$.
    Then $\CE(f_\theta(x_i^\text{adv}(\hat{\alpha})),\tilde{y}_i)\approx \CE(f_\theta(\widehat{x_i}^\text{pgd}),\tilde{y}_i)$.
\end{enumerate}
In fact, \cref{fig:asm2} shows the distribution of the loss ratio, i.e., $\CE(f_\theta(x_i^\text{adv}(\hat{\alpha})),\tilde{y}_i)/\CE(f_\theta(\widehat{x_i}^\text{pgd}),\tilde{y}_i)$, for all training data in three different datasets. It can be observed that the loss ratios mostly concentrate around 1, which empirically supports this assumption. 
Assumption 2 justifies that, for the purpose of adversarial examples, the interpolated adversarial example $x_i^\text{adv}(\hat{\alpha})$ is as effective as an adversarial example generated by PGD with perturbation strength of $\hat{\epsilon}$. Therefore, the linear interpolation between $x_i$ and $x_i^\text{pgd}$ can be used to search for an adversarial example with appropriate perturbation strength with no need to run additional PGD.

\begin{algorithm}
\caption{Semi-Supervised Adversarial Training with Margin-Based Interpolation (SSAT-MBI)}
\label{algo:train}
\begin{algorithmic}[1]
\Statex \textbf{Input:} Labeled dataset $D_L$, unlabeled dataset $D_U$, batch size $B$, margin threshold $\rho$, softmax temperature $\tau$, interpolation steps $K$, outer minimization loss $L$
\Statex \textbf{Output:} Robust classifier $f_\theta$
\State Train a teacher model $f_{\theta_T}$ with a semi-supervised training on $D_L\cup D_U$
\State Assign pseudo-labels $\Tilde{y}_i=\sigma[f_{\theta_T}(x_i)]$ for all $x_i\in D_U$
\State Get fully labeled dataset $\Tilde{D}=D_L\cup \Tilde{D}_U$
\While {training not converged}
    \State Update $\epsilon_\text{max}$ by global epsilon scheduling (\cref{sec:dynamic-epsilon})
    \For {each batch $\{x_i,\Tilde{y}_i\}_{i=1}^B\subset\Tilde{D}$}
        \For {$i \gets 1$ to $B$} 
            \State \label{algline:pgd} $x_i^\text{pgd}=\text{PGD}(\theta, x_i, \Tilde{y}_i, \epsilon_\text{max})$
            \State $\alpha_l=0, \alpha_r=1$
            \For {$\text{step} \gets 1$ to $K$}
                \State $\alpha = (\alpha_l + \alpha_r)/2$
                \State $x_i'=\alpha\cdot x_i^\text{pgd}+(1-\alpha)\cdot x_i$
                \State \label{algline:score} $s=\sigma[f_\theta(x_i')/\tau]$
                \State $d=-\sum_{j=1}^C\Tilde{y}_{i,j}\cdot s_j+\max_{k\in\calY} s_{k}$
                \If {$d<\rho$} $\alpha_l=\alpha$
                \Else $\text{  }\alpha_r=\alpha$
                \EndIf
            \EndFor
            \State $\hat{\alpha}=\alpha_r$
            \State $x_i^\text{adv}=\hat{\alpha}\cdot x_i^\text{pgd}+(1-\hat{\alpha})\cdot x_i$
        \EndFor
\State Update $\theta$ by $\frac{1}{B}\sum_{i=1}^B\nabla_\theta L(\theta,x_i, x_i^\text{adv}, x_i^\text{pgd}, \tilde{y}_i)$
\EndFor
\EndWhile
\State \Return $f_\theta$
\end{algorithmic}
\end{algorithm}

\subsection{Global Epsilon Scheduling}
\label{sec:dynamic-epsilon}
\begin{figure}[t]
  \centering
   \includegraphics[width=0.8\linewidth]{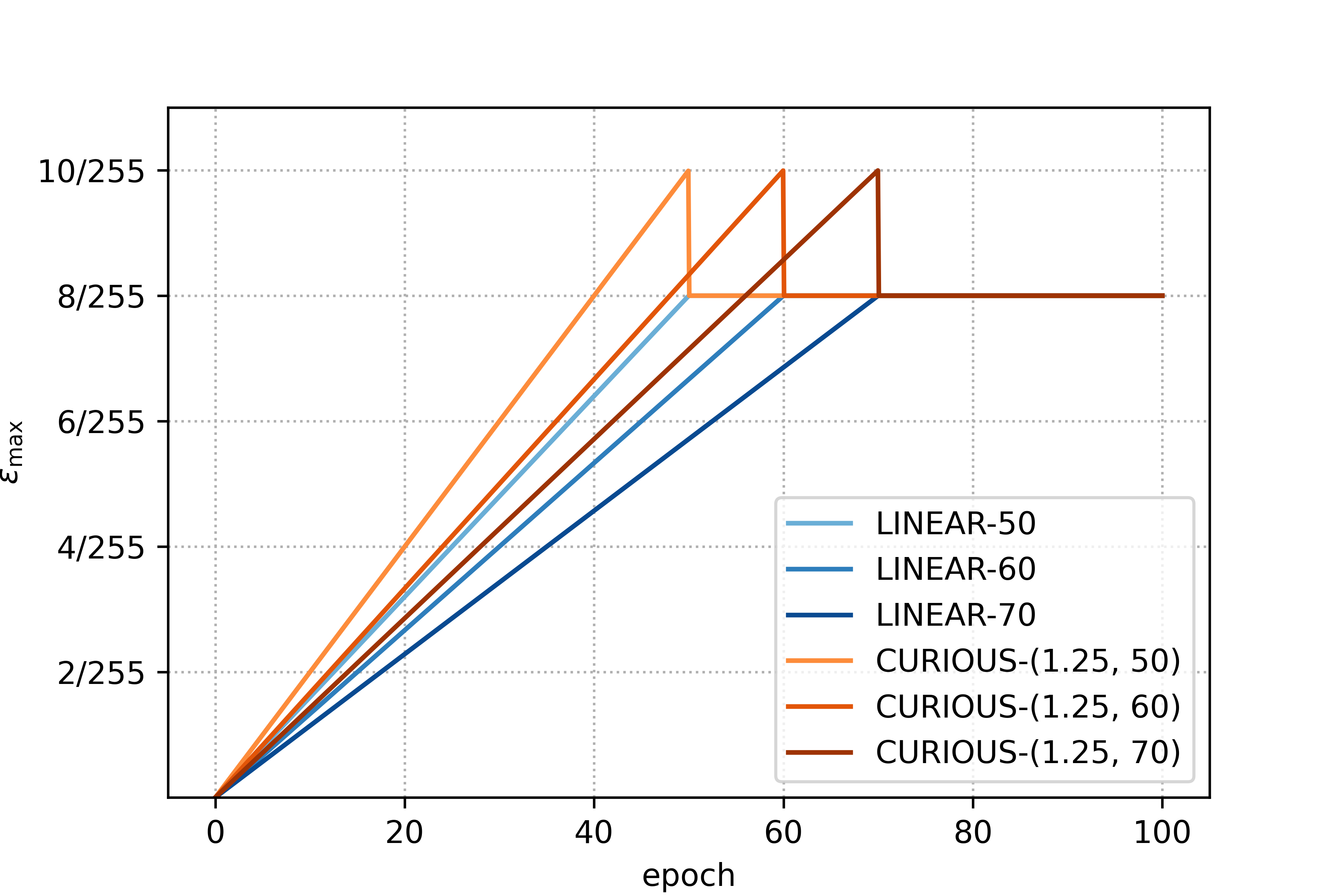}
   \caption{Global epsilon scheduling strategies.}
   \label{fig:eps}
\end{figure}

In~\cref{sec:intp}, an adversarial example $x_i^\text{pgd}$ is generated with the upper bound perturbation strength $\epsilon_\text{max}$ by PGD before interpolation. Using a constant $\epsilon_\text{max}$ (e.g., 8/255) throughout the training can be suboptimal: 
At initial epochs when the model has no robustness against the weakest perturbations, $\epsilon_\text{max}$ may be too aggressive, causing unnecessary distortion in the decision boundaries. At late epochs, $\epsilon_\text{max}$ may be too conservative, limiting further robustness gains.
Therefore, we propose \textbf{global epsilon scheduling}, which dynamically adjusts $\epsilon_\text{max}$ throughout the training, as shown in Line 5 of~\cref{algo:train}.
Considering that prior works~\cite{pgd, trades,rst,uat,srstawr} typically use a fixed $\epsilon=8/255$ under the $\ell_\infty$-norm for adversarial training, 
\cref{fig:eps} shows example global epsilon scheduling strategies:
\begin{itemize}
    \item \textsc{Linear-$t$}: $\epsilon_\text{max}$ linearly increases from $0$ to $8/255$ over the first $t$ epochs and maintains since then. It allows the model to gradually adjust to stronger perturbations and provides a more stable optimization. 
    \item \textsc{Curious-$(\gamma,t)$}: With the parameter $\gamma>1$, $\epsilon_\text{max}$ is linearly increased until $\gamma\cdot\frac{8}{255}$ at epoch $t$, and then settled at $8/255$ afterwards. The temporary exposure to stronger perturbations may encourage learning of slightly more robust features to achieve better robustness against unseen attacks. This can be particularly helpful to explore the vulnerability of hard-to-attack data points. The final settlement at $8/255$ allows the model to finetune its robustness at the conventional level. When $\gamma=1$, it is equivalent to \textsc{Linear-$t$}.
\end{itemize}

Note that while both sample-aware perturbation strength (\cref{sec:saps}) and global epsilon scheduling (\cref{sec:dynamic-epsilon}) adjust the perturbation strength of adversarial examples, they have different effects. 
On one hand, global espilon scheduling controls the upper bound $\epsilon_\text{max}$ of the perturbation strengths for all data points, guiding the adversarial examples to be conservative at initial epochs and be stronger at later epochs. This allows the model to start with learning simple robust features, gradually proceed to complicated ones, and explore stronger ones.
On the other hand, the sample-aware perturbation strength considers each individual data points and controls the margins by which adversarial examples cross the decision boundaries. This balances the contributions across data points so that the model can effectively learn useful information from them.
Our experiments in \cref{sec:ablation} demonstrate that the integration of both methods result in superior robustness and natural accuracy.

\subsection{Semi-Supervised Adversarial Training with Margin-Based Interpolation (SSAT-MBI)}
\label{sec:half}

With the interpolated adversarial examples (\cref{sec:saps}) and global epsilon scheduling (\cref{sec:dynamic-epsilon}), we propose Semi-Supervised Adversarial Training with Margin-Based Interpolation (\textbf{SSAT-MBI}), which is summarized in~\cref{algo:train}. 

In the algorithm, the outer minimization loss $L$ is
\begin{equation}
\label{eq:loss}
\begin{aligned}
L(\theta,x_i,x_i^\text{adv},&x_i^\text{pgd}, \tilde{y}_i)= \CE(f_\theta(x_i),\tilde{y}_i)\\
+\lambda\cdot[\beta\cdot &\KL(p_\theta(\cdot|x_i)\parallel p_\theta(\cdot|x_i^\text{adv})) \\
    +(1-&\beta)\cdot \KL(p_\theta(\cdot|x_i)\parallel p_\theta(\cdot|x_i^\text{pgd}))]
\end{aligned}
\end{equation}
where $\beta\in[0,1]$ is a hyperparameter that balances the contribution from the interpolated adversarial example $x_i^\text{adv}$ and adversarial example $x_i^\text{pgd}$ generated by PGD. In particular, as $x_i^\text{adv}$ is tailored to the decision boundary, it contributes more to the natural accuracy and robustness against PGD attacks, while $x_i^\text{pgd}$ has a stronger perturbation strength, which contributes more to the robustness against stronger attacks like AutoAttack~\cite{autoattack}. 

The above techniques contribute to finding appropriate adversarial examples at the inner maximization phase. As such, our method can also be integrated with existing works that focus on designing outer minimization objectives other than~\cref{eq:loss}, such as SRST-AWR~\cite{srstawr} discussed in~\cref{sec:related}. For instance, we can incorporate our approach with SRST-AWR by introducing the weight factor $w_\theta(x_i)$ in~\cref{eq:srst2} to the KL-Divergence terms in~\cref{eq:loss}. We call this variant \textbf{SSAT-MBI-AWR}, which will be evaluated in~\cref{sec:experiments}.

\begin{table*}[h!]
\centering
\caption{\textbf{Natural accuracy (\%) and robustness of semi-supervised adversarial training methods} on CIFAR-10 (WideResNet-28-5), SVHN (WideResNet-28-2), and CIFAR-100 (WideResNet-28-8). 
Robustness is measured by the accuracy (\%) under the attacks of PGD-10, PGD-20, PGD-40, and AutoAttack, with $\epsilon=8/255$ in the $\ell_\infty$-norm. 
All experiments are conducted 3 times with different random seeds. The reported results are mean values, with standard deviations shown in parentheses.
}
\begin{tabular}{cccccc}
\toprule
\multirow{2}{*}{\textbf{Method}} & \multicolumn{5}{c}{CIFAR-10 (WideResNet-28-5)} \\ 
& \textbf{Clean} & \textbf{PGD-10} & \textbf{PGD-20} & \textbf{PGD-40} & \textbf{AutoAttack} \\ 
\midrule
FixMatch & 95.08 & 0.01 & 0.00 & 0.00 & 0.00\\ 
\midrule
UAT++ & \textbf{84.51} (0.43) & 63.13 (0.23) & 53.67 (0.25) & 52.08 (0.40) & 49.33 (0.34)\\ 
RST  & 82.65 (0.18) & 63.72 (0.14) & 55.94 (0.12) & 54.77 (0.20) & 50.68 (0.13) \\ 
\textbf{SSAT-MBI} & 83.31 (0.25) & \textbf{64.54} (0.13) & \textbf{56.72} (0.02) & \textbf{55.39} (0.10) & \textbf{51.21} (0.15)\\ 
\midrule
SRST-AWR & 84.21 (0.29) & 63.98 (0.12) & 55.35 (0.14) & 54.18 (0.09) &  51.67 (0.18)\\ 
\textbf{SSAT-MBI-AWR} & \textbf{84.61} (0.01) & \textbf{64.42} (0.15) & \textbf{55.92} (0.18) & \textbf{54.95} (0.06) & \textbf{52.00} (0.08)\\

\bottomrule
\toprule
\multirow{2}{*}{\textbf{Method}} & \multicolumn{5}{c}{SVHN (WideResNet-28-2)} \\ 
& \textbf{Clean} & \textbf{PGD-10} & \textbf{PGD-20} & \textbf{PGD-40} & \textbf{AutoAttack}  \\ 
\midrule
FixMatch & 97.40 & 6.70 & 1.18 & 0.32 & 0.00\\ 
\midrule
UAT++ & 90.44 (0.73) & 66.23 (0.92) & 55.04 (1.04) & 52.67 (1.02) & 45.39 (1.20)\\ 
RST  & 91.07 (0.36) & 69.19 (0.29) & 59.92 (0.29) & 58.39 (0.22) & 49.12 (0.10) \\ 
\textbf{SSAT-MBI} & \textbf{91.40} (0.13) & \textbf{69.68} (0.12) & \textbf{60.50} (0.43) & \textbf{59.19} (0.43) &\textbf{49.83} (0.07) \\ 
\midrule
SRST-AWR & 90.98 (0.58) & 69.41 (1.89) & 59.92 (2.36) & 56.92 (2.92) & 45.00 (4.16) \\ 
\textbf{SSAT-MBI-AWR} & \textbf{93.68} (0.13) & \textbf{72.46} (0.46) & \textbf{62.31} (0.60) & \textbf{60.35} (0.57) & \textbf{53.21} (0.11) \\
\bottomrule
\toprule
\multirow{2}{*}{\textbf{Method}} & \multicolumn{5}{c}{CIFAR-100 (WideResNet-28-8)} \\ 
& \textbf{Clean} & \textbf{PGD-10} & \textbf{PGD-20} & \textbf{PGD-40} & \textbf{AutoAttack} \\ 
\midrule
FixMatch & 59.72 & 0.34 & 0.04 & 0.03 & 0.00 \\ 
\midrule
UAT++  & \textbf{51.72} (0.24) & \textbf{35.83} (0.08) & 30.89 (0.10) & 30.34 (0.14) & 25.75 (0.21) \\ 
RST  & 50.60 (0.17) & 35.52 (0.05) & 30.89 (0.23) & 30.49 (0.19) & 26.18 (0.16) \\ 
\textbf{SSAT-MBI} & 50.65 (0.28) & 35.68 (0.20) & \textbf{31.22} (0.14) & \textbf{30.84} (0.16) &\textbf{26.39} (0.06) \\

\midrule
SRST-AWR & 53.45 (0.19) & 36.18 (0.23) & 30.83 (0.25) & 30.28 (0.14) & 25.10 (0.04)\\ 
\textbf{SSAT-MBI-AWR} & \textbf{53.99} (0.38) & \textbf{36.61} (0.23) & \textbf{31.29} (0.19) & \textbf{30.89} (0.18) & \textbf{25.79} (0.08) \\
\bottomrule
\end{tabular}
\label{table:accuracy}
\end{table*}

\subsection{Comparison with Prior Works}
\label{sec:compare-with-prior}
\textbf{Sample-Aware Perturbation Strength: }
Previous works, such as Customized AT~\cite{cat}, MMA~\cite{mma}, FAT~\cite{fat}, and DAAT~\cite{daat}, have also explored adjusting the hardness of adversarial examples for individual data points in adversarial training. However, our approach has key differences. Specifically, Customized AT and MMA generate adversarial examples that merely cross the decision boundaries without controlling the margin. While their methods enhance robustness against PGD attacks, these adversarial examples may be overly tailored to PGD and too weak to ensure robustness against stronger attacks like AutoAttack~\cite{autoattack}, which is demonstrated in the experimental results in the supplementary material.
In contrast, our method identifies the strongest perturbation where the margin of adversarial examples remains within a threshold $\rho$. This ensures that adversarial examples cross the decision boundary by a sufficient margin, enabling the model to learn features that stronger attacks might exploit.
Although FAT and DAAT have similar objectives as our~\cref{eq:aps} in finding adversarial examples, FAT does not explicitly evaluate the margin, and DAAT uses margins in the current epoch to estimate $\epsilon_i$ for next epoch. Both methods can potentially lead to less precise margin control. Experiments that compare their performance with ours are in the supplementary material.

\textbf{Curriculum Learning: }
Previous works such as CAT~\cite{currat} and FAT~\cite{fat} employ curriculum learning strategies that gradually increase the hardness of adversarial examples by increasing the number of PGD steps. 
While our global epsilon scheduling shares the principle of progressing from easier to harder adversarial examples, we take a different approach.
Instead of adjusting the number of PGD steps, we gradually increase the global upper bound of $\epsilon$. 
Moreover, our \textsc{Curious-$(\gamma,t)$} strategy (\cref{sec:dynamic-epsilon}) explores adversarial examples with perturbation strengths greater than $8/255$ during training, which can enhance robustness against strong attacks like AutoAttack.

Besides, all previous works mentioned above focus on fully labeled datasets, whereas our approach is explicitly designed for low-label regimes where labeled training data is scarce and unlabeled data dominates. By effectively leveraging both labeled and unlabeled data, our method is better suited to real-world scenarios where obtaining large labeled datasets is often challenging.

\bigskip

\section{Experiments}
\label{sec:experiments}

\subsection{Experimental Setup}
\label{sec:setup}
Our experiments were conducted on three benchmark datasets: CIFAR-10~\cite{cifar}, SVHN~\cite{svhn}, and CIFAR-100~\cite{cifar}. To simulate the semi-supervised setting, for both CIFAR-10 and CIFAR-100, we retained 4000 labeled samples (8\% of all training samples), and for SVHN, we retained 1000 labeled samples (1.365\% of all training samples). All training data were augmented by random cropping to $32\times32$ and horizontal flipping with a probability of 0.5. We employed Wide Residual Networks (WideResNet)~\cite{wideresnet} as the model architecture. We used WideResNet-28-5 for CIFAR-10, WideResNet-28-8 for CIFAR-100, and WideResNet-28-2 for SVHN. To generate pseudo-labels for the unlabeled data, we leveraged the FixMatch~\cite{fixmatch}, a semi-supervised learning algorithm, to train a teacher model. All models were optimized using Stochastic Gradient Descent. For the generation of adversarial examples during training before interpolation, we employed PGD-10 under the $\ell_\infty$-norm with a perturbation strength of $\epsilon_\text{max}$ and step size set to $\epsilon_\text{max}/4$. The value of $\epsilon_\text{max}$ was determined by the global epsilon scheduling in Section~\ref{sec:dynamic-epsilon}. Further details for the training setup are provided in the supplementary material.

For evaluation, we measured natural accuracy on clean data and robust accuracy against PGD-10, PGD-20, and PGD-40, each representing an iterative attack with a different number of steps. Additionally, we evaluated robust accuracy against AutoAttack~\cite{autoattack}, an ensemble of four advanced adversarial attacks: APGD-CE~\cite{autoattack}, APGD-DLR~\cite{autoattack}, FAB~\cite{fab}, and Square Attack~\cite{square}. 
By evaluating both PGD and AutoAttack, the robustness is validated across diverse adversarial conditions: PGD provides an efficient measure of robustness under practical, time- and resource-constrained attack scenarios, while AutoAttack provides a more rigorous but computationally intensive benchmark. All attacks (except for Square Attack in AutoAttack) are white-box attacks conducted with \(\epsilon=8/255\) under the \(\ell_\infty\)-norm.

\subsection{Performance Evaluation}
\label{sec:main-performance}

To validate the effectiveness of our approach, we compared it with three semi-supervised adversarial training algorithms: UAT++~\cite{uat}, RST~\cite{rst}, and SRST-AWR~\cite{srstawr}. 
We implemented our proposed SSAT-MBI and SSAT-MBI-AWR described in~\cref{sec:half}.
In addition, we also evaluated the performance of models trained by FixMatch~\cite{fixmatch} to show the natural accuracy and robust accuracy of a non-robust model. 

The results in Table~\ref{table:accuracy} demonstrate that our method achieves better robustness under PGD attacks with different steps (PGD-10, PGD-20, and PGD-40) and AutoAttack with $\epsilon=8/255$. Notably, most of the performance improvements exceed the standard deviation range, indicating statistically significant gains in robustness. Although our method sometimes leads to a slight reduction in natural accuracy on clean test data, this trade-off is offset by improvements in robustness across various adversarial attacks. These results highlight the effectiveness of our approach in enhancing the robustness of semi-supervised adversarial learning. 

\subsection{Ablation Studies}
\label{sec:ablation}

\begin{table}[t]
\centering
\caption{\textbf{Ablation study on CIFAR-10 using WideResNet-28-5.} ``Glo." denotes the global epsilon scheduling: ``Cur." refers to \textsc{Curious-$(1.25,70)$}, ``Lin." refers to \textsc{Linear-$60$}, and ``\xmark" means no global epsilon scheduling. ``Int." specifies whether interpolated adversarial examples are used (\cmark) or not (\xmark). All experiments are conducted 3 times with different random seeds. The reported results are mean values, with standard deviations shown in parentheses. }
\begin{tabular}{ccccc}
\toprule
\textbf{Glo.} &  \textbf{Int.} & \textbf{Clean} & \textbf{PGD-20} & \textbf{AutoAttack}\\ 
\midrule
\xmark & \xmark & 82.65 (0.18) & 55.94 (0.12) & 50.68 (0.13) \\ 
Lin. & \xmark & 83.28 (0.09) & 56.29 (0.19) & 50.66 (0.05)\\
Cur. & \xmark  & 82.54 (0.18) & 56.40 (0.09) & 50.85 (0.05) \\ 
\xmark & \cmark & 82.78 (0.25) & 56.02 (0.20) & 50.90 (0.22)\\
Lin. & \cmark & \textbf{83.85} (0.31) & 56.34 (0.14) & 50.52 (0.21)\\
Cur. & \cmark & 83.31 (0.25) & \textbf{56.72} (0.02) & \textbf{51.21} (0.15)\\
\bottomrule
\end{tabular}
\label{table:ablation}
\end{table}

\begin{figure}[tp]
    \centering
    \includegraphics[width=\linewidth]{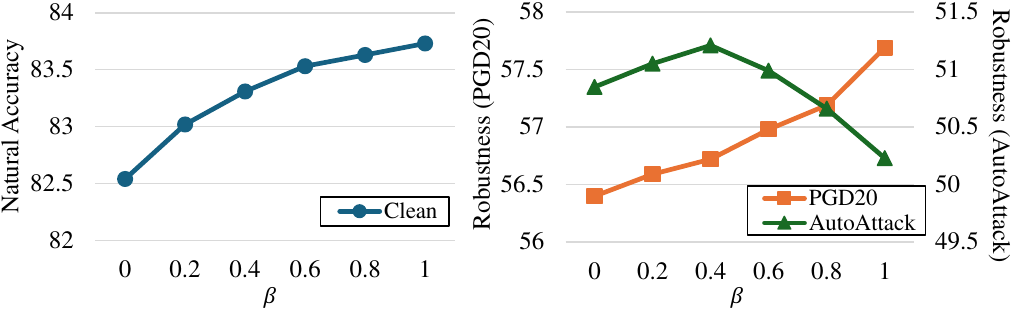}
    \caption{\textbf{Performance of SSAT-MBI with varying $\beta$ on CIFAR-10 using WideResNet-28-5.} The performance is measured by natural accuracy and robustness against PGD-20 and AutoAttack.}
    \label{fig:beta}
\end{figure}

To study the effects of the two techniques proposed in this work, i.e., interpolated adversarial examples and global epsilon scheduling, we conducted ablation studies through experiments of SSAT-MBI on CIFAR-10 as shown in~\cref{table:ablation}. Besides, we 
also study the impact of the hyperparameter $\beta$ introduced in~\cref{sec:half} on model performance. Additional analysis is provided in the supplementary material.

\textbf{Interpolated Adversarial Examples }
\Cref{table:ablation} shows that using interpolated adversarial examples consistently improves robustness across all global epsilon scheduling strategies. This indicates that they can enhance adversarial robustness without compromising the natural accuracy and achieve a balanced performance improvement.

\textbf{Global Epsilon Scheduling}
\Cref{table:ablation} shows that applying \textsc{Linear-$60$} improves natural accuracy and robustness against PGD compared to cases with no global epsilon strategy. However, it slightly reduces the robustness against AutoAttack due to weaker perturbation strengths in adversarial examples. 
In contrast, \textsc{Curious-$(1.25,70)$} enhances natural accuracy and robustness against both PGD and AutoAttack. More experiments to compare different strategies for global epsilon scheduling are shown in the supplementary material.

\textbf{Hyperparameter $\beta$} In~\cref{fig:beta}, we present the performance of SSAT-MBI across different values of the hyperparameter $\beta$. The results indicate that increasing $\beta$ enhances both natural accuracy and robustness against PGD-20, while the highest robustness against AutoAttack is achieved at $\beta=0.4$.

\bigskip

\section{Conclusion}
\label{sec:conclusion}

In this paper, we proposed a semi-supervised adversarial training method called SSAT-MBI to train robust models in low-label regimes where labeled training data is scarce. Our approach focuses on the inner maximization phase by leveraging interpolated adversarial examples that cross decision boundaries by a controlled margin. By accounting for the diversity of data points, our method effectively adjusts the perturbation strength based on each individual data point, allowing the model to learn more informative robust features from the dataset.
Furthermore, our method has a global epsilon scheduling strategy, which determines the global upper bound of perturbation strengths at each epoch. This progressive adjustment allows the model to gradually adapt to more challenging adversarial examples and better learn robust features over time. 
The experiments demonstrated that our method outperforms existing approaches in both natural accuracy and robustness against adversarial attacks.

{\small
\bibliographystyle{ieeenat_fullname}
\bibliography{11_references}
}

\clearpage
\appendix
\section{Additional Details on Experimental Setup}
\label{appendix:setup}

\subsection{Teacher Model}
For all semi-supervised adversarial training experiments in~\cref{sec:main-performance}, a teacher model was trained using FixMatch~\cite{fixmatch}, which optimizes $\theta_T$ with the objective of $l=l_s+\lambda_u\cdot l_u$ consisting of a supervised loss
\[
l_s=\frac{1}{n}\sum_{i=1}^n\CE(f_{\theta_T}(\alpha(x_i)),y_i),
\]
and unsupervised loss
\[
l_u=\frac{1}{m}\sum_{i=n+1}^{n+m} \mathds{1}[\max_yp_{\theta_T}(y|\alpha(x_i))\ge t]\cdot\CE(f_{\theta_T}(\calA(x_i)),\hat{y}_i).
\]
Here, $\hat{y}_i=\arg\max_yp_{\theta_T}(y|\alpha(x_i))$, $t\in(0,1)$ is a constant, and $\alpha$ and $\calA$ are a weak and a strong augmentation, respectively.  More details about FixMatch can be found in~\cite{fixmatch}.

For all three datasets (CIFAR-10, SVHN, CIFAR-100), we followed the same hyperparameter setting for FixMatch as in~\cite{srstawr}. Specifically, $\lambda_u$ is 1, weight decay is $5\times10^{-4}$, $t$ is 0.95, and batch size is 64 for labeled data and 128 for unlabeled data. The performance of the teacher models trained by FixMatch can be found in~\cref{table:accuracy}.

\subsection{Details of SRST-AWR and SSAT-MBI-AWR}

As described in~\cite{srstawr}, SRST-AWR optimizes the following outer minimization objective:
\begin{equation*}
\label{eq:rst}
\begin{aligned}
&\frac{1}{n}\sum_{i=1}^n\ell_{\alpha'}^\text{LS}(f_\theta(x_i),y_i)\\
+&\gamma'\cdot\frac{1}{m}\sum_{j=1}^m\KL(p_{\theta_T}^{\tau'}(\cdot|x_j)\parallel p_{\theta}^{\tau'}(\cdot|x_j))\\
+&\lambda'\cdot\frac{1}{m}\sum_{j=1}^m\KL(p_\theta(\cdot|x_j)\parallel p_\theta(\cdot|x_j^\text{pgd}))\cdot w_\theta(x_j)
\end{aligned}
\end{equation*}
where $\ell_{\alpha'}^\text{LS}$ is the cross-entropy loss with a label smoothing parameter $\alpha'$, and \[
w_\theta(x_j)=\frac{1}{2}\sum_{c=1}^C \tilde{y}_{j,c}p_\theta(c|x_j)+\frac{1}{2}\sum_{c=1}^C\tilde{y}_{j,c}(1-p_\theta(c|x_j^\text{pgd})).
\]
Here, $\tilde{y}_{j,c}=p_{\theta_T}(c|x_j)$ is the pseudo-label generated by the teacher model $\theta_T$, and $x_j^\text{pgd}$ is the adversarial example generated by PGD.  To avoid symbol conflicts, we rename the hyperparameters introduced in~\cite{srstawr} as $\gamma'$, $\lambda'$, and $\tau'$ to distinguish them from similarly named symbols used in our work.

As discussed in~\cref{sec:half}, as our proposed techniques focus on finding appropriate adversarial examples at the inner maximization phase, we can apply our techniques to SRST-AWR, which is called SSAT-MBI-AWR. The outer minimization objective of SSAT-MBI-AWR is shown as follows: 

\begin{equation*}
\label{eq:rst}
\begin{aligned}
&\frac{1}{n}\sum_{i=1}^n\ell_{\alpha'}^\text{LS}(f_\theta(x_i),y_i)\\
+&\gamma'\cdot\frac{1}{m}\sum_{j=1}^m\KL(p_{\theta_T}^{\tau'}(\cdot|x_j)\parallel p_{\theta}^{\tau'}(\cdot|x_j))\\
+&\lambda'\cdot\left[\beta\cdot\frac{1}{m}\sum_{j=1}^m\KL(p_\theta(\cdot|x_j)\parallel p_\theta(\cdot|x_j^\text{adv}))\cdot w_\theta^\text{adv}(x_j)\right.\\
&+\left.(1-\beta)\cdot\frac{1}{m}\sum_{j=1}^m\KL(p_\theta(\cdot|x_j)\parallel p_\theta(\cdot|x_j^\text{pgd}))\cdot w_\theta(x_j)\right]
\end{aligned}
\end{equation*}
where 
\[
w_\theta^\text{adv}(x_j)=\frac{1}{2}\sum_{c=1}^C \tilde{y}_{j,c} p_\theta(c|x_j)+\frac{1}{2}\sum_{c=1}^C\tilde{y}_{j,c}(1-p_\theta(c|x_j^\text{adv})),
\]
and
\[
w_\theta(x_j)=\frac{1}{2}\sum_{c=1}^C \tilde{y}_{j,c}p_\theta(c|x_j)+\frac{1}{2}\sum_{c=1}^C\tilde{y}_{j,c}(1-p_\theta(c|x_j^\text{pgd})).
\]
Here, $x_j^\text{pgd}$ is the adversarial example of $x_j$ generated by PGD, and $x_j^\text{adv}$ is the interpolated adversarial example that we proposed in~\cref{sec:saps}. Note that $\beta\in[0,1]$ is the hyperparameter we introduced in~\cref{sec:half}.

\subsection{Additional Details for Training Setup in~\cref{sec:setup}}
\textbf{For UAT++, RST, and SSAT-MBI: }
Models are optimized using SGD with the initial learning rate of 0.1, momentum of 0.9, and weight decay of $2\times10^{-4}$. The models are trained for 100 epochs, with the learning rate decayed by a factor of 0.1 at epoch 60, by 0.01 at epoch 70, and by 0.005 at epoch 90. A batch size of 128 is used during training. 

\textbf{For SRST-AWR and SSAT-MBI-AWR: }
The training setup follows the original paper~\cite{srstawr}: The SGD optimizer is used with the initial learning rate of 0.1, momentum of 0.9, and weight decay of $5\times10^{-4}$. For our SSAT-MBI-AWR on CIFAR-10 and SVHN, the models are trained for 120 epochs, with the learning rate decayed by a factor of 0.1 at epoch 50. For other cases, the models are trained for 200 epochs, with the learning rate decayed by a factor of 0.1 at epoch 50 and epoch 150. The batch size is 64 for labeled data and 128 for unlabeled data. After 50 epochs, stochastic weighting average (SWA) is used. All models are evaluated using the checkpoint with the highest validation accuracy PGD-10.

\textbf{Other hyperparameters: }
\Cref{table:setup} shows the hyperparameters we used in our experiments for~\cref{sec:main-performance}. The hyperparameters for UAT++ and RST are based on the paper~\cite{srstawr} but slightly tuned for better robustness. The hyperparameters for SRST-AWR follow their original paper~\cite{srstawr}. Note that, for SRST-AWR, there are a few more hyperparameters that we did not mention when we introduce SRST-AWR in~\cref{sec:related}. The setting for these hyperparameters completely follows the paper~\cite{srstawr} and is omitted in~\cref{table:setup}.

\section{Additional Analysis}
\label{appendix:experiments}

\subsection{Effect of Global Epsilon Scheduling Strategies}
\label{appendix:ges}

We conducted experiments to evaluate the impact of different global epsilon scheduling strategies. For \textsc{Linear-$t$}, we tested $t\in\{50, 60, 70\}$. For \textsc{Curious-$(\gamma,t)$}, we tested $\gamma\in\{1.25, 1.5\}$ and $t\in\{50, 60, 70\}$. We also included \textsc{Const}, which represents the case when no global epsilon scheduling is applied, i.e., $\epsilon_\text{max}$ is always $8/255$. The experiments were conducted on CIFAR-10 dataset with 8\% labeled data and 92\% unlabeled data, using WideResNet-28-5 architecture. For all experiments, the hyperparameters are set in the same way as SSAT-MBI, except using different global epsilon scheduling strategies. Note that interpolated adversarial examples are enabled for all experiments.

As shown in~\cref{table:ges}, \textsc{Linear-$50$}, \textsc{Linear-$60$}, and \textsc{Linear-$70$} improve the natural accuracy on clean data and robustness against PGD-20 compared to \textsc{Const}, because they allow the model to learn from adversarial examples with weaker perturbation strengths before $\epsilon_\text{max}$ increases to $8/255$. However, due to the same reason, they show slightly worse robustness against AutoAttack. This drawback is mitigated by \textsc{Curious-$(1.25,50)$},  \textsc{Curious-$(1.25,60)$}, and \textsc{Curious-$(1.25,70)$}, which temporarily increase $\epsilon_\text{max}$ beyond $8/255$ up to $10/255$ before settling it back to $8/255$ at the $50$th, $60$th, and $70$th epoch, respectively. Compared to \textsc{Const}, they effectively improve the robustness of models, especially in the case of \textsc{Curious-$(1.25,70)$} where the best robustness against PGD-20 and AutoAttack is achieved. As a trade-off, their natural accuracy slightly get lower than \textsc{Linear} (still outperform \textsc{Const}), which is because adversarial examples get farther from clean data points. This degradation is more obvious in \textsc{Curious-$(1.5,50)$}, \textsc{Curious-$(1.5,60)$}, and \textsc{Curious-$(1.5,70)$}, when the maximum $\epsilon_\text{max}$ is further increased to $12/255$. This shows the necessity of selecting appropriate $\gamma$ to achieve a good trade-off between natural accuracy and robustness.

\subsection{Ablation Studies for SSAT-MBI-AWR}

In~\Cref{table:ablation-awr}, we present ablation studies for the effects of interpolated adversarial examples and global epsilon scheduling through experiments of SSAT-MBI-AWR on CIFAR-10. The results show that interpolated adversarial examples consistently enhance both natural accuracy and robustness. For global epsilon scheduling, \textsc{Linear}-$60$ improves natural accuracy and robustness against both PGD-20 and AutoAttack, while \textsc{Curious}-$(1.25,60)$ further enhances robustness against AutoAttack, with a slight trade-off in natural accuracy and robustness against PGD-20. These observations align with the studies in~\cref{sec:ablation}, further validating the effectiveness of our proposed techniques. 

\setcounter{table}{2}
\begin{table}[t]
\centering
\caption{Performance of different global epsilon scheduling strategies on CIFAR-10 using WideResNet-28-5. All experiments have interpolated adversarial examples enabled. Natural accuracy and robustness against PGD-20 and AutoAttack are reported. All experiments are conducted 3 times with different random seeds.
The reported results are mean values, with standard deviations shown in parentheses. The best performance for each metric is highlighted in \textbf{bold}, and the second-best is \underline{underlined}.}
\setlength{\tabcolsep}{4.5pt}
\begin{tabular}{@{}cccc@{}}
\toprule
\textbf{Strategy} & \textbf{Clean} &  \textbf{PGD-20} &\textbf{AutoAttack}\\ 
\midrule
\textsc{Const} & 82.78 (0.25) & 56.02 (0.20) & 50.90 (0.22) \\
\textsc{Linear-$50$} & 83.71 (0.24) & 56.36 (0.15) & 50.70 (0.46) \\
\textsc{Linear-$60$} & \underline{83.85} (0.31) & 56.34 (0.14) & 50.52 (0.21) \\
\textsc{Linear-$70$} & \textbf{84.52} (0.32) & 56.38 (0.02) & 50.51 (0.20) \\
\textsc{Curious-$(1.25,50)$}  &  83.09 (0.28) & 56.15 (0.08) & 50.49 (0.20) \\
\textsc{Curious-$(1.25,60)$}  & 83.39 (0.32) & 56.42 (0.17) & 51.00 (0.13) \\
\textsc{Curious-$(1.25,70)$}  & 83.31 (0.25) & \underline{56.72} (0.02) & \underline{51.21} (0.15)\\
\textsc{Curious-$(1.5,50)$}  &  82.87 (0.30) & 56.44 (0.23) & 50.82 (0.15)\\
\textsc{Curious-$(1.5,60)$}  &  82.62 (0.29) & 56.20 (0.17) & 50.91 (0.26) \\
\textsc{Curious-$(1.5,70)$}  &  82.42 (0.23) & \textbf{56.76} (0.12)& \textbf{51.41} (0.16)  \\\bottomrule
\end{tabular}
\label{table:ges}
\end{table}

\subsection{Training Time for Experiments in~\cref{sec:main-performance}}
In~\Cref{table:time}, we present the average training time of our algorithm compared to the baselines. Since not all experiments use the same number of epochs, we also report the training time per epoch. The results demonstrate that our use of interpolated adversarial examples incurs only a minor computational overhead, which supports the claim we made in~\cref{sec:intp}. 

\begin{table}[t]
\centering
\caption{\textbf{Ablation study of SSAT-MBI-AWR on CIFAR-10 using WideResNet-28-5.} ``Glo." denotes the global epsilon scheduling: ``Cur." refers to \textsc{Curious-$(1.25,60)$}, ``Lin." refers to \textsc{Linear-$60$}, and ``\xmark" means no global epsilon scheduling. ``Int." specifies whether interpolated adversarial examples are used (\cmark) or not (\xmark). All experiments are conducted 3 times with different random seeds. The reported results are mean values, with standard deviations shown in parentheses. The best performance for each metric is highlighted in \textbf{bold}, and the second-best is \underline{underlined}.}
\begin{tabular}{ccccc}
\toprule
\textbf{Glo.} &  \textbf{Int.} & \textbf{Clean} & \textbf{PGD-20} & \textbf{AutoAttack}\\ 
\midrule
\xmark & \xmark & 84.21 (0.29) & 55.35 (0.14) & 51.67 (0.18) \\ 
Lin. & \xmark & 84.36 (0.13) & 55.93 (0.10) & 51.86 (0.28)\\
Cur. & \xmark  & 83.18 (0.26) & 55.69 (0.22) & 51.77 (0.16) \\ 
\xmark & \cmark & 84.35 (0.11) & 55.63 (0.17) & 51.68 (0.14)\\
Lin. & \cmark & \textbf{85.46} (0.05) & \textbf{56.02} (0.09) & \underline{51.87} (0.11)\\
Cur. & \cmark & \underline{84.61} (0.01) & \underline{55.92} (0.18) & \textbf{52.00} (0.08)\\
\bottomrule
\end{tabular}
\label{table:ablation-awr}
\end{table}

\begin{figure*}[htbp]
  \centering
    \begin{subfigure}[b]{0.33\textwidth}
    \centering
    \includegraphics[height=4.3cm]{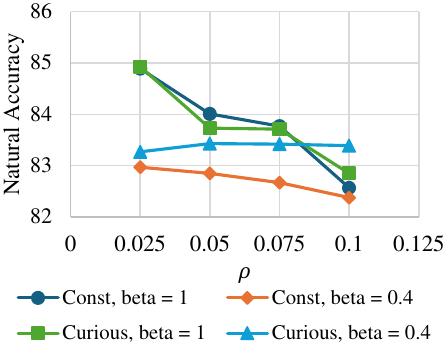}
    \label{fig:rho-clean}
    \end{subfigure}
    \hfill
    \begin{subfigure}[b]{0.33\textwidth}
    \centering
    \includegraphics[height=4.3cm]{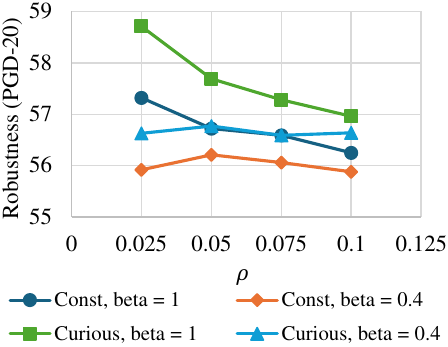} 
    \label{fig:rho-pgd}
    \end{subfigure}
    \hfill
    \begin{subfigure}[b]{0.33\textwidth}
    \centering
    \includegraphics[height=4.3cm]{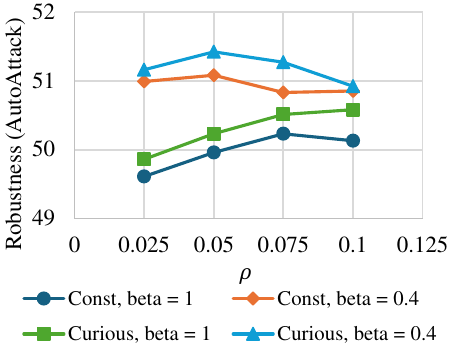}
    \label{fig:rho-aa}
    \end{subfigure}
  \caption{\textbf{Performance of SSAT-MBI with varying $\rho$ on CIFAR-10 using WideResNet-28-5.} The performance is measured by natural accuracy and robustness against PGD-20 and AutoAttack. ``Curious" refers to global epsilon scheduling \textsc{Curious}-$(1.25, 70)$. ``Const" means no global epsilon scheduling. ``beta" refers to the hyperparameter $\beta$ introduced in~\cref{sec:half}.}
  \label{fig:rho}
\end{figure*}

\subsection{Sensitivity Analysis on $\rho$}

We performed a sensitivity analysis on the hyperparameter $\rho$ which controls the margin between interpolated adversarial examples and decision boundaries. In this analysis, $\rho$ represents its initial value at epoch 1, which will be doubled at epoch 75.
We varied the initial $\rho$ from 0.025 to 0.1 and evaluate performance across four settings: $\beta=0.4$ and $1$, each with and without applying \textsc{Curious}-$(1.25,70)$ for global epsilon scheduling.

\Cref{fig:rho} demonstrates that, When $\beta=1$, increasing $\rho$ enhances robustness against AutoAttack but reduces natural accuracy and robustness against PGD-20. This occurs because a higher $\rho$ pushes the interpolated adversarial examples farther from the decision boundaries, leading to more aggressive perturbations during training. These aggressive perturbations strengthen the model’s robustness against strong attacks like AutoAttack but come at the cost of natural accuracy and weaker robustness against PGD-20. 

In contrast, when $\beta=0.4$, the robustness against both PGD-20 and AutoAttack is the highest at $\rho=0.05$, and a higher $\rho$ does not result in better robustness. This can be explained by taking the effect of $x^\text{pgd}$ into account. As $\beta<1$, both $x^\text{pgd}$ and $x^\text{adv}$ are utilized to train the model. Since $x^\text{pgd}$ is generated by PGD without interpolation, it exploits the vulnerability of the model using stronger perturbations. 
For optimal performance, $\rho$ should be reasonably small to balance the contributions from $x^\text{pgd}$ and $x^\text{adv}$, allowing the model to effectively leverage these perturbations to improve robustness.

Additionally, for the same $\beta$ and $\rho$, applying \textsc{Curious}-$(1.25,70)$ consistently results in better performance compared to not applying it. This further demonstrates the effectiveness of our global epsilon scheduling approach.

\begin{table*}[h!]
\centering
\caption{\textbf{Average training time of semi-supervised adversarial training methods} on CIFAR-10 (WideResNet-28-5), SVHN (WideResNet-28-2), and CIFAR-100 (WideResNet-28-8). Total training time is reported in hours (hrs). Training time per epoch is reported in minutes (min).
}
\begin{tabular}{c@{}ccc}
\toprule
\multirow{2}{*}{\textbf{Method}} & \multicolumn{3}{c}{CIFAR-10 (WideResNet-28-5)} \\ 
& \textbf{Epochs} & \textbf{Time (hrs)} & \textbf{Time/epoch (min)} \\ 
\midrule
UAT++ & 100 & 4.62 & 2.77 \\ 
RST & 100 & 4.82 & 2.89 \\ 
\textbf{SSAT-MBI} & 100 & 5.10 & 3.06 \\ 
\midrule
SRST-AWR & 200 & 11.93 & 3.58\\ 
\textbf{SSAT-MBI-AWR} & 120 & 7.73 & 3.87\\
\bottomrule
\toprule
\multirow{2}{*}{\textbf{Method}} & \multicolumn{3}{c}{SVHN (WideResNet-28-2)} \\ 
& \textbf{Epochs} & \textbf{Time (hrs)} & \textbf{Time/epoch (min)} \\ 
\midrule
UAT++ & 100 & 3.08 & 1.85 \\ 
RST & 100 & 3.12 & 1.87 \\ 
\textbf{SSAT-MBI} & 100 & 3.33 & 2.00\\ 
\midrule
SRST-AWR & 200 & 14.12 & 4.24\\ 
\textbf{SSAT-MBI-AWR} & 120 & 8.90 & 4.45\\ 
\bottomrule
\toprule
\multirow{2}{*}{\textbf{Method}} & \multicolumn{3}{c}{CIFAR-100 (WideResNet-28-8)} \\ 
& \textbf{Epochs} & \textbf{Time (hrs)} & \textbf{Time/epoch (min)} \\ 
\midrule
UAT++ & 100 &8.67 &5.20 \\ 
RST  & 100& 8.75& 5.25\\ 
\textbf{SSAT-MBI} & 100 & 9.40 & 5.64\\ 
\midrule
SRST-AWR & 200 & 13.67 & 4.10 \\ 
\textbf{SSAT-MBI-AWR} & 200 & 14.84 & 4.45\\ 
\bottomrule
\end{tabular}
\label{table:time}
\end{table*}

\begin{table}[t]
\centering
\caption{Performance with fully labeled data of CIFAR-10 using WideResNet-28-5.}
\begin{tabular}{cccc}
\toprule
\textbf{Method} & \textbf{Clean} &  \textbf{PGD-20} &  \textbf{AutoAttack}\\ 
\midrule
Customized AT~\cite{cat} & 93.31 & 66.11 & 16.76 \\
MMA~\cite{mma} & 87.40 & 54.50 & 42.36 \\
\midrule
TRADES~\cite{trades} & 82.59 & 55.50 & 50.63 \\
TRADES+FAT~\cite{fat} & 84.23 & 55.21 & 50.70 \\
\textbf{TRADES+Ours} & \textbf{84.65} & \textbf{56.86} & \textbf{50.99} \\

\bottomrule
\end{tabular}
\label{table:sup}
\end{table}

\begin{table}[t]
\centering
\begin{threeparttable}
\caption{Performance with fully labeled data of CIFAR-10 using ResNet18.}
\begin{tabular}{cccc}
\toprule
\textbf{Method} & \textbf{Clean} &  \textbf{PGD-20} &  \textbf{AutoAttack}\\ 
\midrule
TRADES & 81.98\tnote{*} & 53.70\tnote{*} & 49.22\tnote{*} \\
TRADES+DAAT~\cite{daat} & \textbf{83.55}\tnote{*} & 54.57\tnote{*} & 49.83\tnote{*} \\
\textbf{TRADES+Ours} & 83.44 & \textbf{56.71} & \textbf{49.92} \\

\bottomrule
\end{tabular}
\begin{tablenotes}
\footnotesize
\item[*] Results are taken directly from the original paper of DAAT~\cite{daat} due to the unavailability of its open-source code.
\end{tablenotes}
\label{table:sup-daat}
\end{threeparttable}
\end{table}

\subsection{Comparison with Supervised Adversarial Training Methods}
\label{sec:sup}
We evaluated our method on the fully labeled CIFAR-10 dataset and compared it with several supervised adversarial training methods that share similar concepts to our approach, as discussed in~\cref{sec:compare-with-prior}. The comparison includes Customized AT\footnote{\url{https://github.com/hirokiadachi/Customized-Adversarial-Training}}~\cite{cat}, MMA\footnote{\url{https://github.com/BorealisAI/mma_training}}~\cite{mma}, and FAT\footnote{\url{https://github.com/zjfheart/Friendly-Adversarial-Training}}~\cite{fat}, which have publicly available open-source codes.
Additionally, we included TRADES\footnote{\url{https://github.com/yaodongyu/TRADES}}~\cite{trades} as a baseline, which is a widely used supervised adversarial training method known for effectively balancing robustness and natural accuracy. For FAT, we evaluated the version integrated with TRADES, which is the most effective version as reported in their paper~\cite{fat}. We implemented the supervised version of our method by applying interpolated adversarial examples with $\rho=0.05$ and $K=3$, and \textsc{Curious}-$(1.25,70)$ on top of TRADES. All experiments utilized the WideResNet-28-5 architecture. 

As shown in~\cref{table:sup}, while Customized AT~\cite{cat} demonstrates high robustness against PGD-20 attacks, and both Customized AT and MMA~\cite{mma} achieve high natural accuracy, they exhibit significantly lower robustness against AutoAttack. 
This may be attributed to the absence of a threshold $\rho$ to control the minimum margin between their adversarial examples and the decision boundaries. Without such a mechanism, the adversarial examples can be too weak to ensure robustness against strong attacks, as discussed in~\cref{sec:compare-with-prior}.
Compared to TRADES~\cite{trades}, FAT shows improved natural accuracy and slightly higher robustness against AutoAttack. We our method is applied to TRADES, it results in a more substantial improvement in natural accuracy and robustness against both PGD-20 and AutoAttack. 

We also compared our method with DAAT~\cite{daat}. As their open-source code is unavailable, we report the results of DAAT directly from their original paper. Following their experimental setup, we used ResNet18 as the model architecture. Our method is integrated with TRADES, with $\rho=0.05$, $K=3$ and \textsc{Curious}-$(1.25,70)$. As shown in~\cref{table:sup-daat}, our method outperforms DAAT in robustness against PGD-20 with a good margin and against AutoAttack, while achieving a comparable  natural accuracy.

\begin{table*}
\centering
\caption{Hyperparameters for experiments in~\cref{table:accuracy}. $\lambda$ is as defined in~\cref{eq:rst,eq:uat}. $K$ is the number of steps for binary search. $\tau$ is the temperature as defined in \cref{eq:score}. $\beta$ is as defined in \cref{sec:half}. ``Inner Max." represents the loss function used for inner maximization, which can be either cross-entropy (CE) or Kullback–Leibler divergence (KL). ``Global Eps. Sch." represents the strategy for global epsilon scheduling. $\rho$ is the margin threshold.}
\begin{tabular}{ccccccccc}
\toprule
\textbf{Dataset} & \textbf{Method} & $\lambda$ & $K$ & $\tau$ & $\beta$ & \textbf{Inner Max.} & \textbf{Global Eps. Sch.} & $\rho$ \\
\midrule
\multirow{7}{*}{CIFAR-10} 
 & UAT++ & 8 & / & / & / & CE & / & / \\
 & RST & 8 & / & / & / & CE & / & / \\
 & \multirow{2}{*}{\textbf{SSAT-MBI}} & \multirow{2}{*}{8} & \multirow{2}{*}{3} & \multirow{2}{*}{2} & \multirow{2}{*}{0.4} & \multirow{2}{*}{CE} & \multirow{2}{*}{\textsc{Curious-$(1.25,70)$}} & \{0.05, 0.1\} \\
 & & & & & & & & at epoch \{1, 75\} \\
 & SRST-AWR & 20  & / & / & / & KL & / & / \\
 & \multirow{2}{*}{\textbf{SSAT-MBI-AWR}} & \multirow{2}{*}{20} & \multirow{2}{*}{3} & \multirow{2}{*}{2} & \multirow{2}{*}{0.5} & \multirow{2}{*}{KL} & \multirow{2}{*}{\textsc{Curious-$(1.25,60)$}} & \{0.05, 0.1\} \\
 & & & & & & & & at epoch \{1, 75\} \\
\midrule
\multirow{5}{*}{SVHN} 
 & UAT++ & 4 &  / & / & / & CE & / & / \\
 & RST & 4 &  / & / & / & CE & / & / \\
 & \textbf{SSAT-MBI} & 4 & 3 & 2 & 0.5 & CE & \textsc{Curious-$(1.25,60)$} & 0.05 \\
 & SRST-AWR & 15 & / & / & / & CE & / & / \\
 & \textbf{SSAT-MBI-AWR} & 15 & 3 & 2 & 0.5 & CE & \textsc{Curious-$(1.25,60)$} & 0.05 \\
\midrule
\multirow{5}{*}{CIFAR-100} 
 & UAT++ & 6  & / & / & / & CE & / & / \\
 & RST & 6  & / & / & / & CE &/ & / \\
 & \textbf{SSAT-MBI} & 6 & 4 & 2 & 1.0 & CE & \textsc{Curious-$(1.25,60)$} & 0.05 \\
 & SRST-AWR & 20 & / & / & / & CE & / & / \\
 & \textbf{SSAT-MBI-AWR} & 20 & 3 & 2 & 1.0 & CE & \textsc{Curious-$(1.25,60)$} & 0.05 \\
\bottomrule
\end{tabular}
\label{table:setup}
\end{table*} 

\end{document}